# Ensemble learning of diffractive optical networks

Md Sadman Sakib Rahman[1,2,3,†], Jingxi Li[1,2,3,†], Deniz Mengu[1,2,3], Yair Rivenson[1,2,3] and Aydogan Ozcan[1,2,3,*]

[1]Electrical and Computer Engineering Department, University of California, Los Angeles, CA, 90095, USA

[2]Bioengineering Department, University of California, Los Angeles, CA, 90095, USA

[3]California NanoSystems Institute (CNSI), University of California, Los Angeles, CA, 90095, USA

[*]Correspondence to: ozcan@ucla.edu

[†]These authors contributed equally to this work.

## Abstract

A plethora of research advances have emerged in the fields of optics and photonics that benefit from harnessing the power of machine learning. Specifically, there has been a revival of interest in optical computing hardware, due to its potential advantages for machine learning tasks in terms of parallelization, power efficiency and computation speed. Diffractive Deep Neural Networks ($D^2$NNs) form such an optical computing framework, which benefits from deep learning-based design of successive diffractive layers to all-optically process information as the input light diffracts through these passive layers. $D^2$NNs have demonstrated success in various tasks, including e.g., object classification, spectral-encoding of information, optical pulse shaping and imaging, among others. Here, we significantly improve the inference performance of diffractive optical networks using feature engineering and ensemble learning. After independently training a total of 1252 $D^2$NNs that were diversely engineered with a variety of passive input filters, we applied a pruning algorithm to select an optimized ensemble of $D^2$NNs that collectively improve their image classification accuracy. Through this pruning, we numerically demonstrated that ensembles of N=14 and N=30 $D^2$NNs achieve blind testing accuracies of 61.14±0.23% and 62.13±0.05%, respectively, on the classification of CIFAR-10 test images, providing an inference improvement of >16% compared to the average performance of the individual $D^2$NNs within each ensemble. These results constitute the highest inference accuracies achieved to date by any diffractive optical neural network design on the same dataset and might provide a significant leapfrog to extend the application space of diffractive optical image classification and machine vision systems.

**Introduction**

Recent years have witnessed the emergence of deep learning[1], which has facilitated powerful solutions to an array of intricate problems in artificial intelligence, including e.g., image classification[2,3], object detection[4], natural language processing[5], speech processing[6], bioinformatics[7], optical microscopy[8,9], holography[10–12], sensing[13] and many more[14]. Deep learning has become particularly popular because of the recent advances in the development of advanced computing hardware and the availability of large amounts of data for training of deep neural networks. Algorithms such as stochastic gradient descent and error back propagation enable deep neural networks to learn the mapping between an input and the target output distribution by processing a large number of examples. Motivated by this major success enabled by deep learning, there has also been a revival of interest in optical computing[15–29], which has some important and appealing features such as e.g., (1) parallelism provided by optics/photonics systems, (2) potentially improved power efficiency through e.g., passive and/or low-loss optical interactions, and (3) minimal latency.

As a recent example of an entirely passive optical computing system, Diffractive Deep Neural Networks ($D^2NN$)[18,24,26,30–35] have been demonstrated to perform all-optical inference and image classification through the modulation of input optical waves by successive diffractive surfaces that are trained through deep learning methods, e.g., stochastic gradient-descent and error-backpropagation. Earlier generations of these diffractive neural networks achieved >98% blind testing accuracies in classification of handwritten digits (MNIST) that are encoded in the amplitude or phase channels of the input optical fields, and were experimentally demonstrated using terahertz wavelengths along with 3D-printing of the resulting diffractive layers/surfaces that form a physical network. In a $D^2NN$ that is fabricated with linear materials, where nonlinear optical processes including surface nonlinearities are negligible, the only form of nonlinearity in the forward optical model occurs at the opto-electronic detector plane. Without the use of any non-linear activation function, $D^2NN$ framework still exhibits depth feature as its statistical inference and generalization capabilities improve with additional diffractive layers, which was demonstrated both empirically[26,36] and theoretically[35]. The same diffractive processing framework of $D^2NNs$ has also been utilized to design deterministic optical components for e.g., ultra-short pulse shaping, spectral filtering and wavelength division multiplexing[31,33].

To further improve the inference capabilities of optical computing hardware, coupling of diffractive optical systems with jointly-trained electronic neural networks that form opto-electronic hybrid systems have also been reported[19,26,30], where the front-end is optical/diffractive and the back-end is all-electronic. Despite all this progress, there is still significant room for further improvements in diffractive processing of optical information.

Here we demonstrate major advances in the optical inference and generalization capabilities of $D^2NN$ framework by feature engineering and ensemble learning over multiple independently trained diffractive neural networks, where we exploit parallel processing of optical information. To create this advancement, first we focused on diversifying the base $D^2NN$ models by manipulating their training inputs by means of spatial feature engineering. In this approach, the input fields are filtered either in the object space or in the Fourier space by introducing an assortment of curated passive filters, before the diffractive networks (see Fig. 1). Following the individual training of 1252 uniquely different $D^2NN$s with various features, we used an iterative pruning strategy to obtain ensembles of $D^2NN$s that work in parallel to improve the final classification accuracy by combining the decisions of the individual diffractive classifiers. Based on this feature learning and iterative pruning strategy, we numerically achieved blind testing accuracies of 61.14±0.23% and 62.13±0.05% on the classification of CIFAR-10 test images with ensemble sizes of N=14 and N=30 $D^2NN$s, respectively. Stated differently, 14 $D^2NN$s (30 $D^2NN$s) selected through this pruning approach work in parallel to collectively reach 61.14±0.23% (62.13±0.05%) optical inference accuracy for CIFAR-10 test dataset, which provides an improvement of >16% over the average classification accuracy of the individual $D^2NN$s within each ensemble, demonstrating the "*wisdom of the crowd*". This image classification performance is the highest achieved to date by any diffractive optical network design, applied on the same dataset. We believe that this significantly improved inference and generalization performance provided by feature engineering and ensemble learning of $D^2NN$s marks a major step forward to open up new avenues for optics-based computation, machine learning and machine vision related systems, benefiting from parallelism of optical systems.

**Results**

Ensemble learning refers to improving the inference capability of a system by training multiple models instead of a single model, and combining the predictions of the constituent models (known as base models, base learners or inducers). It is also possible to learn how to combine the decisions of the base learners, which is known as meta-learning[37] (learning from learners). Ensemble learning is beneficial for several reasons[38]; if the size of the training data is small, the base learners are prone to overfitting and as a result suffer from poor generalization to unseen data. Combining multiple base learners helps to ameliorate this problem. Also, by combining different models, the hypothesis space can be extended and the probability of getting stuck in a local minimum is reduced. An important aspect to consider when generating ensembles is the diversity of the learned base models[38]. The learned models should be diverse enough to ensure that different models learn from different attributes of the data, such that through their "*collective wisdom*" the ensemble of these models can eliminate the implicit variance of the constituent models and substantially improve the collective inference performance. One approach to enrich the diversity of the base models is to manipulate the training data used to train different classifiers, making them learn different features of the input space in each trained model. On top of the training of these unique and independent classifiers, pruning methods that aim at finding small sized ensembles, while also achieving a competitive inference performance are also very important[38].

Based on these considerations, Figs. 1a and 1b depict the two types of $D^2NNs$[30] (base learners) that we have selected to constitute our ensemble diffractive system. The difference between these two types lies in the placement of the input mask (passive) used to filter out different spatial features of the object field to variegate the information fed to the base $D^2NN$ classifiers. In the structure of Fig. 1a, the input filter is placed on the object plane, whereas the structure of Fig. 1b uses an input filter on the Fourier plane of a 4-f system placed before the $D^2NN$. Further heterogeneity is introduced by diversifying the input filter profiles for both types depicted in Figs. 1a and 1b (see the Supplementary Table S1). For example, input filters with transmissive windows of different shapes (rectangular, Gaussian, Hamming, Hanning) and different locations are used at the object plane. The input filters used at the Fourier plane also vary in terms of their pass/stop bands (see the Materials and Methods section for more details). To further improve the diversity of the base models, the input object information is encoded into either the phase channel with 4 different dynamic ranges, or the amplitude channel of the illumination field. Using all of these different hyperparameter choices and their combinations, 1252 unique $D^2NN$ classifiers were trained to form the initial pool of our networks. 340 of these networks had the input object information encoded in the amplitude channel, while 912 of them had phase encoded inputs. 276 of the amplitude encoded $D^2NNs$ had an input filter located on the object plane and 64 had an input filter located on the Fourier plane. 656 of the phase-encoded-input networks had a filter on the object plane and 256 had a filter on the Fourier plane. For these 1252 unique $D^2NN$ classifiers, each diffractive neural network subsequently acts on the filtered version of the input image, and therefore the trained diffractive layers of each base $D^2NN$ directly act on the space domain information (*not* the frequency or Fourier domain).

The preparation of this initial set of 1252 unique $D^2NNs$ was followed by iterative pruning, with the aim of obtaining ensembles of significantly reduced size, i.e., with much smaller number of $D^2NNs$ (base models) in the ensemble. Ensemble pruning was performed by assigning weights to each class score of the individual $D^2NN$ classifiers and defining the ensemble class score as a weighted sum of the individual class scores. At each iteration of the ensemble pruning, the weights were optimized through the gradient descent and error backpropagation method to minimize the softmax-cross-entropy (SCE) loss between the predicted ensemble class scores and their one-hot labeled ground truth, followed by choosing the set of weights giving the highest accuracy (see the Materials and Methods section). Then, the 'significance' of the individual $D^2NNs$ in a given state of the ensemble was quantified and ranked by the absolute summation (i.e., L1 norm) of their weights, based on which a certain fraction of the networks was then eliminated from the ensemble due to their relatively minor contributions. In addition to this greedy search, a periodic *random* elimination of the individual classifiers from the ensemble was also used in the pruning process, so that the solution space could be expanded (see the Materials and Methods section for details).

Based on this pruning process, the iterative search algorithm resulted in a sequence of $D^2NN$ ensembles with gradually decreasing sizes. To select the final ensemble with a desired size (i.e., the number of unique networks), we set a maximum limit on the ensemble size (referred to as the 'maximum allowed ensemble size', i.e., $N_{max}$), and searched for the $D^2NN$ ensemble that achieves the best performance in terms of inference accuracy on the *validation* dataset (i.e., the test dataset was never used during the pruning phase). As we followed this procedure for different values of the pruning hyperparameters, $D^2NN$ ensembles with different sizes and blind testing accuracies were created; we repeated our search 3 times for each set of hyperparameters, which helped us quantify the mean and standard deviation of the inference accuracy for the resulting $D^2NN$ ensembles. Based on these analyses, Fig. 2a reveals that as the maximum allowed ensemble size ($N_{max}$) gets larger, the blind testing accuracies increase; Fig. 2b shows a similar trend reporting the blind testing accuracies as a function of N, i.e., the number of $D^2NNs$ in the ensemble that is selected. Figure 2c further reports the relationship between N and $N_{max}$ during the pruning process, which indicates that on average these two quantities vary linearly (with a slope of ~1).

While the results reported in Figs. 2a,b demonstrate the significant gains achieved through the ensemble learning of diffractive networks, they also highlight a diminishing return on the blind inference accuracy of the ensemble with increasing number of $D^2NNs$ selected. For example, with ensemble sizes of N=14 and N=30 $D^2NNs$, we achieved blind inference, image classification accuracies of 61.14±0.23% and 62.13±0.05%, respectively, on CIFAR-10 test dataset. Increasing the ensemble size to e.g., N=77 $D^2NNs$ resulted in a classification accuracy of 62.56% on the same test dataset. Because of this diminishing return achieved by larger ensemble sizes, we further focused on the case of $N_{max}$=14 to better explore a sweet-spot: Table 1 reports the blind testing accuracies achieved for different pruning hyperparameters for a maximum allowed ensemble size of 14. These results summarized in Table 1 reveal that, although non-intuitive, the periodic *random* elimination of diffractive models during the pruning process results in better classification accuracies, compared to pruning with no random model elimination; see the columns in Table 1 with $T = \infty$, where $T$ refers to the interval between periodic random elimination of $D^2NN$ models. In Table 1, the best average blind testing accuracy (61.14±0.23%) that was achieved for $N_{max}$=14 is highlighted with a green box. For 3 individual repeats of the pruning process using the same hyperparameters, the classification accuracies achieved by the resulting 14 $D^2NNs$ were 60.88%, 61.33% and 61.21%. Figure 3 further presents a detailed analysis of the latter N=14 ensemble that achieved a blind testing accuracy of 61.21%, which is the median among the 3 repeats. Six of the selected base $D^2NN$ classifiers have input filters on the object plane, while the remaining eight have input filters on the Fourier plane (Fig. 3a). Figure 3b also shows the magnitudes of the class specific weights, optimized for the base classifiers of this N=14 ensemble. Even if these optimized weights are ignored, and made all to be equal to 1, the same diffractive ensemble of 14 $D^2NNs$ achieves a similar inference accuracy of 61.08%, a small reduction from 61.21%.

In addition to these, Fig. 3c also shows the true positive rates for each class, corresponding to the individual members of N=14 $D^2NNs$ as well as the ensemble. The improvements in the true positive rates of the ensemble over the mean performance of the individual classifiers for different data classes lie between 13.47% (for class 0) and 19.98% (for class 6). Figure 3d further presents a comparison of the classification accuracies of the individual diffractive classifiers compared against their ensemble. Through these comparative analyses reported in Figs. 3c and 3d, it is evident that the performance of the ensemble is significantly better than any individual diffractive network of the ensemble, demonstrating the "wisdom of the crowd" achieved through our pruning process.

In Table 1, we also report another metric, i.e., 'the accuracy per network', which is the average accuracy divided by the number of networks in the ensemble, to reveal the performance efficiency of ensembles that achieve at least 60% average blind testing accuracy for CIFAR-10 test dataset. The best performance achieved in Table 1 based on this metric is highlighted with a red box: N=12 unique $D^2NN$s selected by the pruning process with $N_{max}$=14 achieved a blind testing accuracy of 60.35±0.39%, where the accuracy values for the individual 3 repeats were 60.77%, 60.00% and 60.29%. Details of the latter ensemble with a blind testing accuracy of 60.29% (which is the median among the 3 repeats) can be found in Supplementary Fig. S1, revealing the selected input filters and the class specific weights of the resulting 12 $D^2NN$s of this ensemble.

Our results reveal that encoding the input object information in the amplitude channel of some of the base $D^2NN$s and in the phase channel of the other $D^2NN$s help to diversify the ensemble. Supplementary Table S2 further confirms this by reporting the blind testing accuracies achieved when the initial ensemble consists of *only* the 912 $D^2NN$s whose input is encoded in the phase channel. A direct comparison of Table 1 and Supplementary Table S2 reveals that including both types of input encoding (phase and amplitude) within the ensemble helps improve the inference accuracy. Using *only* phase encoding for the input of $D^2NN$s, the best average blind testing accuracy achieved using $N_{max}$=14 was 60.74±0.17 with an ensemble size of N=14. The detailed description of the *median* of these $D^2NN$ ensembles with a classification test accuracy of 60.65% is provided in Supplementary Fig. S2. Supplementary Fig. S3 also shows the details of another phase-only input encoding ensemble with N=12 $D^2NN$s, achieving a blind testing accuracy of 60.43%.

Finally, it is noteworthy that the top 10 $D^2NN$s (in terms of their individual blind testing accuracies) within the initial pool of 1252 networks were **not** selected in any of the $D^2NN$ ensembles of Fig. 3 and Supplementary Figs. S1, S2 and S3. *This corroborates our conjecture that the individual performance of a base model might not be indicative of its performance within an ensemble*. In fact, several of the base $D^2NN$s selected in the ensembles of Fig. 3 and Supplementary Figs. S1, S2 and S3 had blind testing accuracies <40%, whereas the blind testing accuracies of the best models (not chosen in any of the ensembles) were >50%.

**Discussion**

Although forming an ensemble of separately trained $D^2NN$s ensues a major improvement in the classification and generalization performance of diffractive networks, further improvements might be possible to reduce the performance gap with respect to the state-of-the-art electronic neural networks. The classification accuracies of widely known all-electronic classifiers on grayscale CIFAR-10 test image dataset can be summarized as[30]: Support Vector Machine (SVM)[39] 37.13%, LeNet[40] 66.43%, AlexNet[2] 72.64%, ResNet[3] 87.54%. While the blind testing accuracy of an ensemble of N=30 unique diffractive optical networks (62.13±0.05%) comes close to the performance of LeNet, which was the first demonstration of a convolutional neural network (CNN), there is still a large performance gap with respect to the state-of-the-art CNNs, and this suggests that there might be more room for improvements, especially through a wider span of input feature engineering within larger pools of $D^2$NNs, forming a much richer and more diverse initial condition for iterative pruning.

The presented improvement in the classification performance of $D^2$NNs obtained with feature engineering and ensemble learning does not come free of cost. Due to the multiple optical paths as part of this framework, the number of diffractive layers and the opto-electronic detectors to be fabricated and used increases in proportion to the number of networks (N) used in the final ensemble, which results in an increased complexity for the optical network set-up. The required training time also raises up significantly because of the need for a large number of individual networks in the initial pool, which in our case was 1252 individual $D^2$NNs. However, this training process is a one-time effort, and the inference time or latency remains the same by virtue of the parallel processing capability of the diffractive optical system; stated differently, the information processing occurs through diffraction of light within each $D^2$NN of the ensemble, and because all of the individual diffractive networks of an ensemble are passive devices that work in parallel, we do not expect a slowdown in speed of inference. Also, the detection circuitry complexity of the diffractive optics based solutions is still minimal compared to its electronic counterparts, and the hardware complexity of $D^2$NN ensembles can be reduced even further by using an additive sum of the individual class scores instead of the weighted sum, at the cost of a very small sacrifice in the inference accuracy. For example, for the ensemble of $D^2$NNs depicted in Fig. 3, if a simple additive sum of the individual class scores is used instead of the optimized class-specific weights, the blind classification accuracy reduces only slightly from 61.21% to 61.08%. This suggests that a further reduction in the hardware complexity is attainable with a very small sacrifice in the inference accuracy by discarding the specific weights of the class scores. However, these weights still play a very significant role in the pruning process as they help in our selection of the diffractive models to be retained in each iteration during the ensemble pruning by measuring/quantifying the significance of the individual networks in an ensemble (see the Materials and Methods section).

Some of the drawbacks associated with the relatively increased size and complexity of the optical hardware should also become less restrictive since the advances in integrated photonics and fabrication technologies have led to continuous miniaturization of opto-electronic devices[41]. In addition to the issues of hardware complexity and size, to maintain a desired signal-to-noise ratio at the output detectors, the optical input (illumination) power of the system needs to be increased in proportion to the ensemble size. However, due to the availability of various high-power laser sources, this higher demand for increased illumination power of the system will not be a significant obstacle for its operation.

In summary, we significantly improved the statistical inference and generalization performance of $D^2NNs$ using feature engineering and ensemble learning. We independently trained a total of 1252 unique $D^2NNs$ that were diversely engineered with various passive input filters. Using a pruning algorithm, we searched through these 1252 $D^2NNs$ to select an ensemble that collectively improves the image classification accuracy of the optical network. Our results revealed that ensembles of N=14 and N=30 $D^2NNs$ achieve blind testing accuracies of 61.14±0.23% and 62.13±0.05%, respectively, on the classification of CIFAR-10 test images, which constitute the highest inference accuracies achieved to date by any diffractive optical neural network design applied on this dataset. The versatility of $D^2NN$ framework stems from its applicability to different parts of the electromagnetic spectrum and the availability of miscellaneous fabrication techniques such as 3D printing and lithography. Together with further advances in the miniaturization and fabrication of optical systems, the presented results and the underlying platform might be utilized in a variety of applications, for e.g. ultra-fast object classification, diffraction-based optical computing hardware, and computational imaging tasks.

**Materials and Methods**
**Implementation of $D^2NNs$.** As the basic building block of our diffractive ensemble, all the individual $D^2NN$ base classifiers presented in this paper consist of 5 successive diffractive layers, which modulate the phase of the incidence optical field and are connected to each other by free space propagation in air. The propagation model we used was formulated based on the Rayleigh-Sommerfeld diffraction equation[18,26], assuming that each diffractive feature (or "neuron") on the diffractive layers serves as a source of modulated secondary waves, which jointly form the propagated wave field. The presented results and analyses of this manuscript are broadly applicable to any part of the electromagnetic spectrum as long as the diffractive features and the physical dimensions are accordingly scaled with respect to the wavelength of light. Using a coherent illumination wavelength of λ, for all the diffractive network designs, the size of each neuron and the axial distance between two successive diffractive layers were set to be ~0.5 λ and 40 λ, respectively, which guarantees an adequate diffraction cone for each neuron to optically communicate with all the neurons of the consecutive layer, and enables the diffractive optical network to be "fully-connected". Each photodetector at the output plane of a $D^2NN$ is assumed to be a square, with a width of 6.4 λ. Since we employed here a differential detection scheme[30], the detectors were divided into two groups: positive detectors and negative detectors, and were collectively used to compute the differential class scores for network $k$, i.e., $z_{ck}$ through the following equation:

$$z_{ck} = \frac{z_{ck,+} - z_{ck,-}}{z_{ck,-} + z_{ck,-}} \quad (1)$$

where $z_{ck,+}$ and $z_{ck,-}$ denote the optical signal of the positive and the negative detectors for class *c*, respectively. An empirical factor of *K=0.1*, also termed as the "temperature" coefficient in machine learning literature[42], was a non-trainable hyperparameter that we utilized to achieve more efficient convergence during the training phase by dividing Eq. (1) by *K*. In addition, the input object was encoded either in the amplitude or in the phase channel of the input illumination, which is assumed to be a uniform plane wave generated by a coherent source. The phase encoding of the input objects took values from either of the following four intervals: 0-0.5π, 0-π, 0-1.5π or 0-2π.

**Feature engineering of diffractive networks.** We used two types of feature engineered diffractive networks: one employed an input filter placed on/against the object plane that filters the spatial signals directly, while the other one used an input filter placed on the Fourier plane of a 4-f system to filter certain spatial frequency components of the object. Unless the filters are specifically mentioned to be trainable, these input filter designs were pre-defined, keeping the transmittance of their pixels *constant* during the training of the diffractive networks (see e.g., Supplementary Table S1). For all the feature engineered D²NN classifiers, each diffractive network subsequently acts on the filtered input image, directly processing the input information on the spatial domain, *not* the frequency or Fourier domain.

The object plane filters are designed to be of the same size as the object, containing transmissive patterns, the amplitude distribution of which takes one of the following forms: 1) 2D Gaussian functions defined with variable shapes and center positions; 2) multiple superposed 2D Gaussian functions defined with variable center positions; 3) 2D Hamming/Hanning functions defined with variable center positions; 4) square windows with different sizes at variable center positions; 5) multiple square windows at variable center positions; 6) patch-shaped windows rotated at variable angles; 7) circular windows at variable center positions; 8) sinusoidal gratings with variable periods and orientations; 9) Fresnel zone plates with variable x-y spatial positions; and 10) superpositions of Gaussian functions and square windows at variable spatial x-y positions.

For the second type of D²NNs with a Fourier plane input filter, using the same Rayleigh-Sommerfeld diffraction equation mentioned above, we numerically implemented a 4-f system with two lenses; the first lens transforms the object information from the spatial domain to the frequency domain and the second one does the opposite. On the Fourier plane that is 2f away from the object plane, a single amplitude-only input filter, designed in one of the following forms is employed: 1) various combinations of circular/annular passbands, which are defined through specifying a series of equally spaced concentric ring-like areas, such that it can serve as a low/high pass, single-band pass or multi-band pass filter, or 2) a single *trainable* layer enabling the system to learn an input spatial frequency filter on its own. On the output image plane of the 4-f system that is 4f away from the object plane, a square aperture is placed with the same size as the object or 1.5 times the size of the object, before feeding the resulting complex-valued field into the diffractive network. In the numerical implementation, the lens has a focal length $f$ of ~145.6 $\lambda$ and a diameter of 104 $\lambda$.

For each type of the input filter design, the number of trained base $D^2NNs$ and some input filter examples can be found in Supplementary Table S1.

**Ensemble pruning.** The method we followed for ensemble pruning involved iterative elimination of the $D^2NN$ members from the initial pool of 1252 unique networks based on a quantitative metric, which is indicative of an individual network's "significance" in the collective inference process. However, since a member's individual performance supremacy might not always translate to an improvement of the ensemble, during the iterative process we occasionally eliminated some members *randomly*. Ensemble pruning with intermittent random elimination of members was found to result in better performing ensembles compared to pruning without random elimination as detailed in the Results section and Table 1.

Our pruning method (see Fig. 4) was initiated with an ensemble that consisted of all the $n_0 = 1252$ individually trained $D^2NN$ models. An ensemble class score $z_c$ was defined as $z_c = \sum_k w_{ck} z_{ck}$, where $z_{ck}$ is the score predicted for class $c$ by the member/network $k$ (Eq. 1), and $w_{ck}$ is the corresponding class-specific weight. The weight vectors $w_k = \{w_{ck}\}_{c=1}^{C}$, $k = 1, 2, …, n_0$, were optimized by minimizing the softmax cross-entropy loss of the class scores predicted by the ensemble of $D^2NNs$; $C=10$ denotes the total number of classes in our dataset. To reduce overfitting the weights to the training data examples, an L2 loss term was also included in our pruning loss function:

$$\text{Pruning loss} = -E\left[\sum_{c=1}^{C} g_c \log\left(\frac{\exp(z_c)}{\sum_{c=1}^{C}\exp(z_c)}\right)\right] + \alpha\left(\frac{1}{2}\sum_{k=1}^{n_0}\sum_{c=1}^{C} w_{ck}^2\right) \quad (2)$$

where $\alpha$ is set to be 0.001, $E[.]$ denotes the expectation over the image batch, and $g_c$ represents the $c^{th}$ entry of the ground truth label vector $g$. During the optimization of the ensemble, in each iteration of the back propagation, all the image samples in the validation set were fed into the ensemble model (i.e. the batch size equals to 5K); using training images for weight optimization during the ensemble pruning resulted in overfitting and therefore was not implemented. The class-specific weights were optimized using the gradient-descent algorithm (Adam[43]) for 10000 steps. After optimizing the weights, the individual members/networks were ranked based on a quantitative metric. An intuitive choice for this metric could have been the individual prediction accuracy of each network. However, a better metric for measuring the significance of individual networks in an ensemble was found to be the L1 norm of the individual weight vectors optimized for validation accuracy. The superiority of the weight L1 norm as a metric was substantiated by the fact that it resulted in ensembles achieving much better blind testing accuracies, consistently.

After ranking the members based on their weight vectors, a certain fraction of them was eliminated from the bottom (i.e., lowest ranked ones), and the procedure was repeated with the reduced ensemble until only one member was left in the ensemble. As mentioned earlier, after *T* iterations of the pruning, this member/network elimination was done *randomly* instead of the ranking based elimination. However, to avoid elimination of the members with the largest weights, random elimination was select within a fraction *p* (which was 2/3 in our case) of the networks counted from the bottom. Once the pruning process was complete (see Fig. 4), a maximum allowed ensemble size ($N_{max}$) was set, and the ensemble with the best performance on the validation dataset and satisfying the size limit was chosen. *The test image dataset was never used during the pruning process*.

**Training details.** All the $D^2NNs$ and their weighted ensembles in this paper were numerically implemented and trained using Python (v3.6.5) and TensorFlow (v1.15.0, Google Inc.). An Adam optimizer[43] with default parameters (predefined in TensorFlow) was used to calculate the back-propagated gradients during the training of the individual optical models and the ensemble weights. The learning rate, starting from an initial value of 0.001, was set to decay at the rate of 0.7 every 8 epochs. Since the images in the original CIFAR-10 dataset contain three color channels (red, green and blue) and monochromatic illumination is used in our diffractive optical network models, the built-in *rgb_to_grayscale* function in TensorFlow was applied to convert these color images to grayscale. Also, to enhance the generalization capability of the trained $D^2NNs$, we randomly flipped the images (left-to-right) with a probability of 0.5 while training. For training of the individual $D^2NNs$, we used a batch size of 8 and trained each model for 50 epochs using the training image set, and selected the best model based on the classification performance on the validation image set. The $D^2NN$ loss function for a given network *k* used softmax-cross-entropy between the differential class scores $z_{ck}$ and their one-hot labeled ground truth vector *g*:

$$D^2NN\ Loss = -E\left[\sum_{c=1}^{C} g_c \log\left(\frac{\exp(z_{ck})}{\sum_{c=1}^{C}\exp(z_{ck})}\right)\right] \quad (3)$$

where $E[.]$ denotes the expectation over the training images in the current batch, *C*=10 denotes the total number of classes in the dataset, and $g_c$ represents the $c^{th}$ entry of the ground truth label vector *g*.

For all the training tasks detailed above, we used a desktop computer with a GTX 1080 Ti graphical processing unit (GPU, Nvidia Inc.), Intel® Core TM i7-8700 central processing unit (CPU, Intel Inc.) and 64 GB of RAM, running Windows 10 operating system (Microsoft Inc.). The typical training time of one $D^2NN$ model is ~4.5 hours. The time required for the iterative ensemble pruning process depends on the pruning hypermeters, varying between 0.75 to 7.5 hours.

**Acknowledgement**

The Ozcan Research Group at UCLA acknowledges the support of Fujikura (Japan).

# Figures and Tables

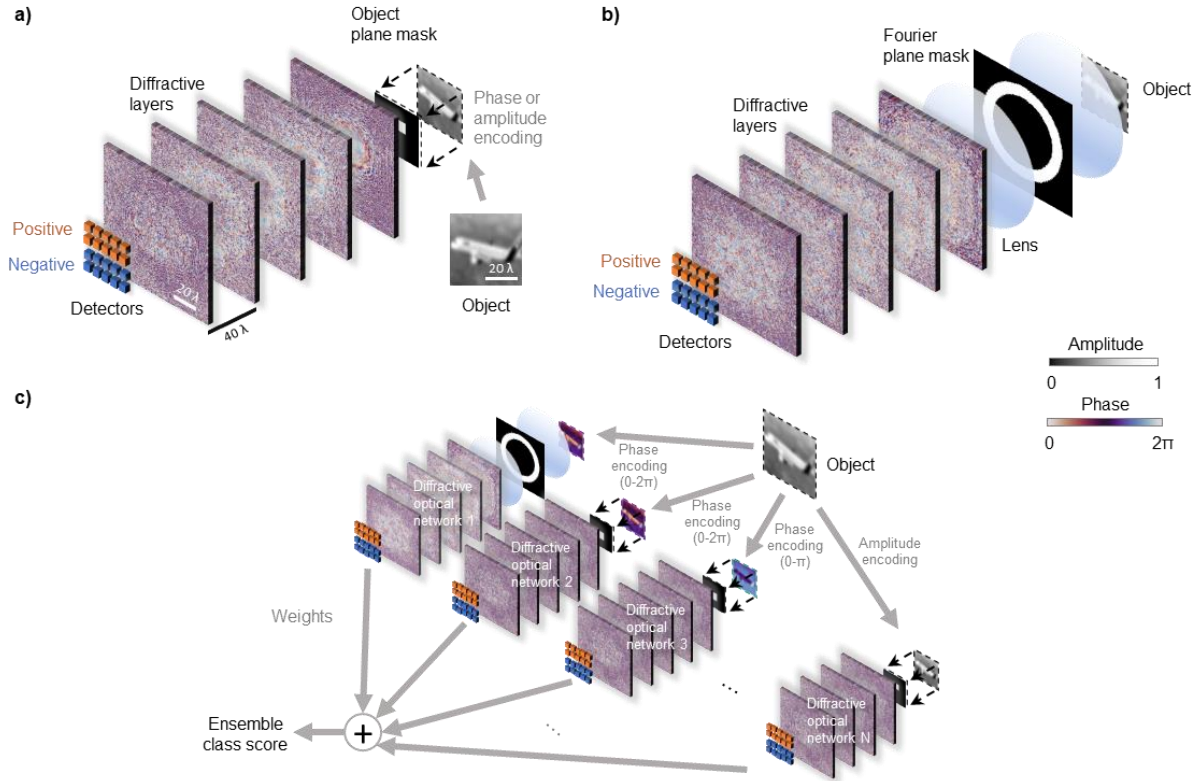

**Fig. 1 Schematic diagram of the ensemble diffractive network system.** (a) Example of a $D^2NN$ using a feature engineered input, where an input mask with a passive transmission window opened at a certain position is employed against the object plane. An object from the CIFAR-10 image dataset is shown as an example and is encoded either in the amplitude channel or in the phase channel of the input plane of the diffractive network. **(b)** Same as in (a), but uses a passive input mask placed on the Fourier plane of a 4-f system; here a bandpass filter is shown as an example. **(c)** An ensemble $D^2NN$ system, formed by N different feature engineered $D^2NNs$, is shown where each diffractive network of the ensemble takes the form of (a) or (b). The final ensemble class score is computed through a weighted summation of the differential detector signals obtained from the individual diffractive networks. Through feature engineering and ensemble learning, we achieved blind inference accuracies of 62.13±0.05%, 61.14±0.23% and 60.35±0.39% on CIFAR-10 test image dataset using N=30, N=14 and N=12 $D^2NNs$, respectively. The standard deviations are calculated through 3 repeats using the same hyperparameters.

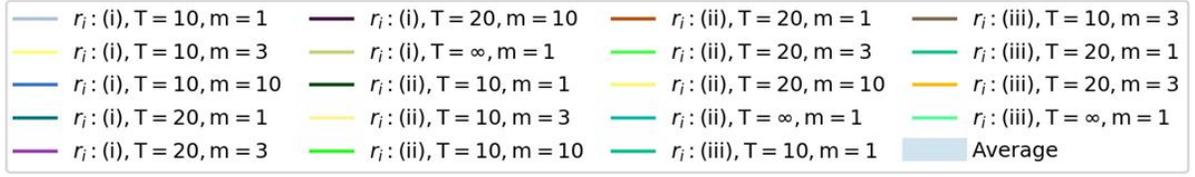
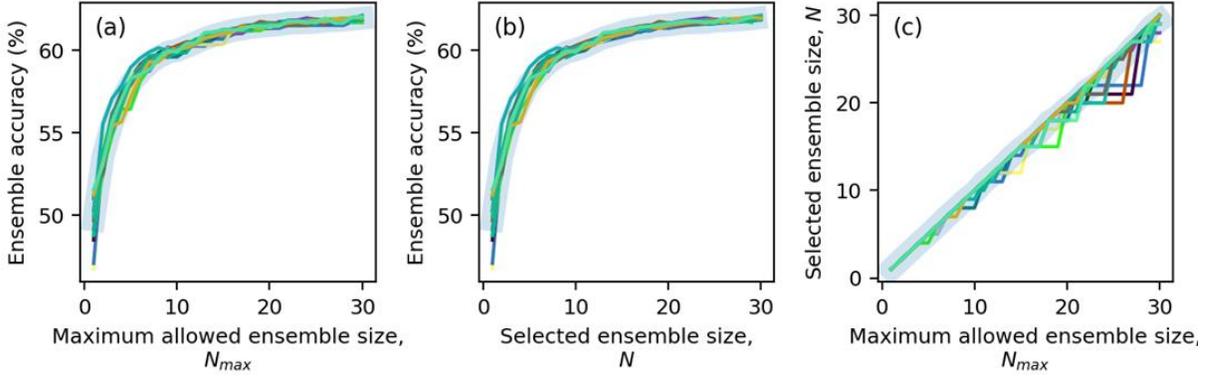

**Fig. 2 Inference accuracy of D$^2$NN ensembles as a function of N$_{max}$ and N.** (a) Variation of the blind testing accuracy as a function of the maximum allowed ensemble size (N$_{max}$) during the pruning; (b) Variation of the blind testing accuracy as a function of the selected ensemble size (N); (c) Relationship between N$_{max}$ and N. The symbols in the legend denote different pruning parameters used in our ensemble selection process; also see Fig. 4 and Table 1.

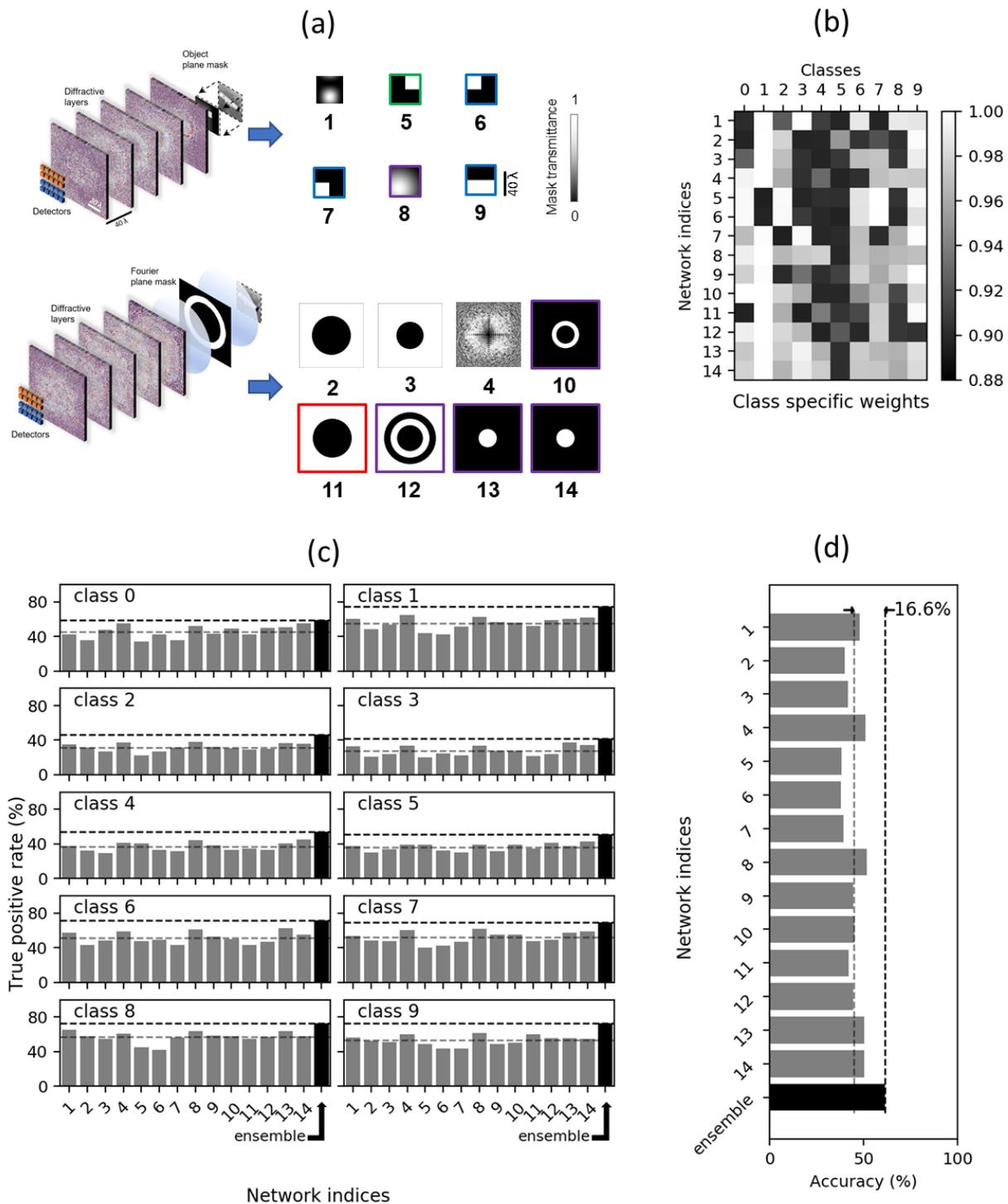

**Fig. 3 An ensemble of N=14 D²NNs achieves a blind classification accuracy of 61.21% on CIFAR-10 test dataset.** (a) Input filters/masks used before each one of the D²NNs that form the ensemble. For D²NNs 1, 5-9: the input filters are on the object plane. For the remaining D²NNs 2-4, 10-14: the input filters are on the Fourier plane. The input filters corresponding to the networks

with phase encoded inputs are enclosed within a border/frame (5-14), while the inputs of the diffractive networks 1-4 are amplitude encoded. The dynamic range of the input phase encoding is represented by the border color; red: $0-\pi/2$, green: $0-\pi$, blue: $0-3\pi/2$, purple: $0-2\pi$. (b) Class specific weights for each $D^2NN$ of the ensemble. If one ignores these class specific weights and replaces them with all ones, the blind inference accuracy slightly decreases to 61.08%, from 61.21%. (c) True positive rates of the individual diffractive networks, compared against their ensemble for different classes. (d) Test accuracy of the individual networks compared against their ensemble. The dotted lines show the classification performance improvement (~16.6%) achieved by the diffractive ensemble over the mean performance of the individual $D^2NNs$. Three repeats with the same hyperparameters resulted in a blind classification accuracy of 61.14±0.23%, where 61.21% represents the median, detailed in this figure.

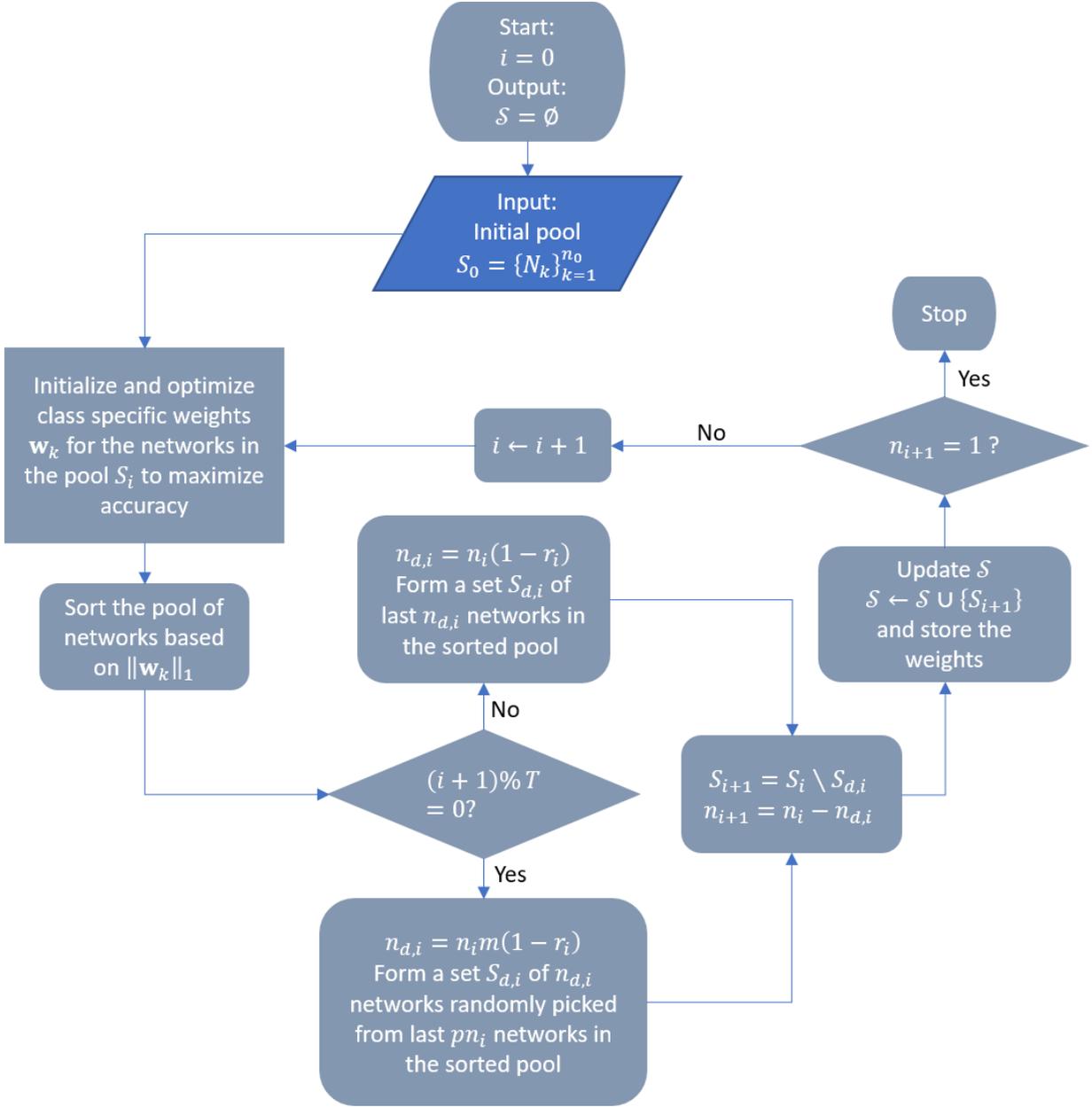

**Fig. 4 Flow chart of the ensemble pruning process.** The meaning of the symbols is as follows: $i$: iteration number. $\mathcal{S}$: the set of ensembles, resulting after each iteration; $S_i$: ensemble on iteration $i$; $n_i$: the number of networks in the ensemble on iteration $i$; $w_k$: the weight vector for network $k$; $T$: the interval between random elimination of D$^2$NNs; $S_{d,i}$: the set of networks to eliminate from the ensemble on iteration $i$; $n_{d,i}$: the number of networks to eliminate from the ensemble on iteration $i$; $r_i$: the fraction of networks to retain on iteration $i$; $m$: the ratio of the number of randomly eliminated networks to the number of networks eliminated based on ranking; $p$: the fraction of the networks in the ensemble on which random elimination is applied. At the end of the pruning process, $\mathcal{S}$ comprises a series of D$^2$NN ensembles (formed by $S_i$) of gradually decreasing size.

| Accuracy (%) <br> Number of networks <br> Accuracy per network(%) | $T$ | 10 | | | 20 | | | $\infty$ | | |
|---|---|---|---|---|---|---|---|---|---|---|
| | $r_i$ | (i) | (ii) | (iii) | (i) | (ii) | (iii) | (i) | (ii) | (iii) |
| $m$ | 1 | 60.813 ± 0.131 | 60.873 ± 0.091 | 60.993 ± 0.195 | 60.857 ± 0.329 | 61.090 ± 0.251 | 61.000 ± 0.277 | 61.120 | 60.830 | 61.120 |
| | | 14 | 14 | 14 | 14 | 14 | 14 | 14 | 14 | 14 |
| | | 4.344 ±009 | 4.348 ±0.006 | 4.357 ±0.014 | 4.347 ±0.023 | 4.364 ±0.018 | 4.357 ±0.020 | 4.366 | 4.345 | 4.366 |
| | 3 | 61.010 ± 0.262 | 60.903 ± 0.261 | 61.140 ± 0.233 | 61.060 ± 0.421 | 60.847 ± 0.142 | 60.887 ± 0.130 | 61.120 | 60.830 | 61.120 |
| | | 14 | 14 | 14 | 14 | 14 | 14 | 14 | 14 | 14 |
| | | 4.358 ±0.019 | 4.350 ±0.019 | 4.367 ±0.017 | 4.361 ±0.030 | 4.346 ±0.010 | 4.349 ±0.009 | 4.366 | 4.345 | 4.366 |
| | 10 | 60.693 ± 0.035 | 61.033 ± 0.152 | | 60.743 ± 0.299 | 60.353 ± 0.389 | | 61.120 | 60.830 | 61.120 |
| | | 14 | 14 | | 14 | 12 | | 14 | 14 | 14 |
| | | 4.335 ±0.003 | 4.360 ±0.011 | | 4.339 ±0.021 | 5.029 ±0.032 | | 4.366 | 4.345 | 4.366 |

| (i) | (ii) | (iii) |
|---|---|---|
| $r_i = 0.98$ | $r_i = 0.98 + (0.9 - 0.98)e^{-i/2}$ | $r_i = \begin{cases} 0.9, i < 20 \\ 0.95, 20 \leq i < 40 \\ 0.98, i \geq 40 \end{cases}$ |

**Table 1. Comparison of blind testing accuracy results achieved under different pruning hyperparameters, with a maximum allowed ensemble size of $N_{max}$=14 (see Fig. 4).** For the classification accuracies that are reported, the average and the standard deviation values result from 3 independent repeats of the pruning process using the same hyperparameters. The lower table describes the schemes used for $r_i$ denoted by (i), (ii) and (iii). The green box highlights the ensemble achieving the best average blind testing accuracy (N=14), and the red box highlights the ensemble achieving the best average blind testing accuracy *per network* (N=12).